\documentclass[10pt,conference]{IEEEtran}
\IEEEoverridecommandlockouts
\usepackage{cite}
\usepackage{amsmath,amssymb,amsfonts}
\usepackage{algorithmic}
\usepackage[dvipdfmx]{graphicx}
\usepackage{textcomp}
\usepackage{xcolor}
\usepackage{booktabs}
\usepackage{multirow}
\usepackage{bm}
\usepackage{bbm}
\usepackage{chngpage}
\usepackage{cellspace}

\def\BibTeX{{\rm B\kern-.05em{\sc i\kern-.025em b}\kern-.08em
    T\kern-.1667em\lower.7ex\hbox{E}\kern-.125emX}}
\begin{document}

\title{LogELECTRA: Self-supervised Anomaly Detection for Unstructured Logs\\
    \thanks{This work was supported by the Cabinet Office (CAO), Cross-ministerial Strategic Innovation Promotion Program (SIP), “Cyber Physical Security for IoT Society” (funding agency: NEDO).}
}

\author{
    \IEEEauthorblockN{Yuuki Yamanaka}
    \IEEEauthorblockA{\textit{NTT Social Informatics Laboratories} \\
        Tokyo, Japan \\
        yuuki.yamanaka@ntt.com}
    \and
    \IEEEauthorblockN{Tomokatsu Takahashi}
    \IEEEauthorblockA{\textit{NTT Social Informatics Laboratories} \\
        Tokyo, Japan \\
        tomokatsu.takahashi@ntt.com}
    \and
    \IEEEauthorblockN{Takuya Minami}
    \IEEEauthorblockA{\textit{NTT Social Informatics Laboratories} \\
        Tokyo, Japan \\
        takuya.minami@ntt.com}
    \and
    \IEEEauthorblockN{Yoshiaki Nakajima}
    \IEEEauthorblockA{\textit{NTT Social Informatics Laboratories} \\
        Tokyo, Japan \\
        yoshiaki.nakajima@ntt.com}
}

\maketitle

\begin{abstract}
    System logs are some of the most important information for the maintenance of software systems, which have become larger and more complex in recent years.
    The goal of log-based anomaly detection is to automatically detect system anomalies by analyzing the large number of logs generated in a short period of time, which is a critical challenge in the real world.
    Previous studies have used a log parser to extract templates from unstructured log data and detect anomalies on the basis of patterns of the template occurrences.
    These methods have limitations for logs with unknown templates.
    Furthermore, since most log anomalies are known to be point anomalies rather than contextual anomalies, detection methods based on occurrence patterns can cause unnecessary delays in detection.
    In this paper, we propose LogELECTRA, a new log anomaly detection model that analyzes a single line of log messages more deeply on the basis of self-supervised anomaly detection.
    LogELECTRA specializes in detecting log anomalies as point anomalies by applying ELECTRA, a natural language processing model, to analyze the semantics of a single line of log messages.
    LogELECTRA outperformed existing state-of-the-art methods in experiments on the public benchmark log datasets BGL, Sprit, and Thunderbird.
\end{abstract}


\section{Introduction}
With advances in the IT industry in recent years, software systems have become larger and more complex, and the difficulty of maintaining them has become a major issue.
System logs, which record the status and events of a system at a certain time stamp, are very important information when maintaining complex software systems.
Through the monitoring of system logs, operators can understand the status of the system and quickly identify problems such as incidents, failures, anomalous behavior and cyber-attacks.
System logs is need to be monitored in real-time to quickly detect anomalies and ensure stable system operation.
However, as software systems become larger, system logs are generated on a large scale in a short period of time, making them difficult to analyze manually in real time.
For this reason, automatic real-time log anomaly detection is being actively researched~\cite{le2022log, chen2021experience, he2016experience, fu2009execution, guo2021logbert, 10.1145/3460345}.

The system log is a sequence of log messages collected simultaneously from the software system.
A log message is a pair of timestamps and a string of characters generated by certain syntax rules, and it contains a record of a specific system event in each line.
The strings in log messages can be further divided into a fixed part (the log template) and a variable part (the log parameters).
For example, the log message "total of 100 MB of files have been downloaded." can be divided into "100 MB" and the rest~\cite{he2017drain, du2016spell, makanju2011lightweight, 4601543, 10.1109/ICSE-SEIP.2019.00021}.

With the recent remarkable progress in the deep learning technology, deep learning is also being used in the log anomaly detection technology.
Existing deep learning-based log anomaly detection techniques have two steps.
The first step is to convert the log message into a feature representation that is vector representation suitable for anomaly detection at a later step.
The conversion to feature representation can be divided into two approaches, depending on the use of a log parser.
A log parser is a tool that can perform parsing on unstructured log messages and the log messages are split into log templates and log parameters.
Typical log parsers include Drain, Spell, IPLoM, and AEL~\cite{he2017drain, du2016spell, makanju2011lightweight, 4601543}.
The time series of log template indices and their frequency vectors are then used as feature representation vectors.
Recently, several methods have been proposed to convert log templates into semantic representation vectors using Word2Vec and BERT models trained on large text corpus such as Wikipedia.
In the parser-free approach, on the other hand, basic preprocessing such as normalization and removal of stop words is performed.
Then, natural language processing models such as Transformer are used to directly embed the text into feature representation vectors.
The second step is anomaly log detection using deep learning.
In this step, the feature representation vectors obtained in the previous step are used to build an anomaly detection model to find anomalous log messages.
Depending on the type of deep learning technique used, it can be divided into supervised and unsupervised approaches.
In the supervised approach, the log anomaly detection problem is formulated as a binary classification problem between normal and abnormal.
The model is typically trained by using cross entropy on the time-series feature vectors of the logs.
On the other hand, the unsupervised approach aims to learn a model describing normal system behavior through logs.
This approach typically formulates the problem as a regression problem to predict the next log from a time series of past logs.
The Recurrent Neural Networks (RNNs) and Transformers are used to learn the time series~\cite{yu2019review, vaswani2017attention}.

Although these existing approaches have shown effectiveness in their respective experiments, they still have problems.
First, the method using a log parser analyzes templates and parameters separately, so some important information is lost in the structuring process.
In particular, it has been pointed out that the methods are unable to correctly detect log messages with unseen log templates that are not included in the training data~\cite{7476654}.
Also, the performance of log anomaly detection is strongly dependent on the performance of the log parser, not on the detection algorithm itself~\cite{le2022log}.
Second, the methods based on supervised learning are not practical.
Supervised learning has the potential to classify normal and abnormal data more accurately than unsupervised learning, but it requires a large number of labeled abnormal data in the training dataset for its training.
However, in real-world software systems, most logs are normal and anomalous logs are rare.
In addition, using anomaly logs for training requires manual labeling of log data by experts, which incurs a significant cost.
Therefore, it is not realistic to train on labeled anomaly data in a real system, and supervised approaches are not suitable for real-world problems~\cite{japkowicz2002class}.
Finally, most unsupervised learning-based methods use time-series analysis of log messages.
Recent studies have shown that almost all anomalous logs can be considered point anomalies that do not require time-series analysis~\cite{10.1007/978-3-031-14135-5_12}.
Therefore, existing methods may examine an unnecessary number of log messages in an online detection setting before each model raises an anomaly alert~\cite{le2022log}.
Early detection of anomalies is important in real systems because it allows more time to consider more mitigating actions.

To solve the aforementioned problem, we propose LogELECTRA, a new log anomaly detection model that analyzes a single line of log messages more deeply on the basis of self-supervised anomaly detection.
LogELECTRA specializes in detecting anomalous logs as point anomalies.
To more accurately capture the characteristics of a single normal log message, LogELECTRA learns the context of the token sequence in a normal log message through an auxiliary task called Token Replacement Detection~\cite{clark2020electra}.
Then, in the evaluation phase, LogELECTRA judges whether the log message has a different context than the learned context and detects anomalies.
Thus, LogELECTRA is robust to unseen log messages because it is parser-free, is practical because it is based on unsupervised approach, and has no detection time lag because it evaluates each line of the log.
LogELECTRA was evaluated against the system log benchmark datasets Blue Gene/L, Spirit and Thunderbird.
The results show that LogELECTRA improves the detection accuracy of anomalous logs compared with various existing methods.

\section{Relatedwork}
In recent years, numerous methods have been proposed to analyze log data and detect anomalous logs using deep learning~\cite{li2020swisslog, guo2021logbert, meng2019loganomaly, du2017deeplog, nedelkoski2020self, yang2021semi, lu2018detecting, zhang2019robust}.
These methods can achieve higher accuracy than conventional data mining based methods~\cite{liang2007failure, xu2009detecting}.
In this section, we briefly introduce deep learning-based methods in the following.

Deeplog~\cite{du2017deeplog} is the first log anomaly detection method using deep learning.
This method uses a log parser to split log messages into log templates and log parameters.
It uses an unsupervised learning-based Long Short-Term Memory (LSTM) model to learn the pattern of normal log sequences in a system by predicting the next log event from a past sequence of log templates~\cite{hochreiter1997long}.
In the evaluation phase, the observed sequence of log messages is input into the model, and if the actual log event is included up to the top $k$-th of the predicted next log event, then the log message is considered normal.

LogAnomaly~\cite{meng2019loganomaly} utilizes an Attention-based Bi-LSTM model to perform unsupervised learning of both frequency and sequence patterns of log templates.
It is a prediction-based anomaly detection model similar to DeepLog, which predicts the next log event and detects anomalies when observed log events differ from the predicted results.
LogAnomaly is a log parser-dependent method, but utilizes the semantics of unseen logs by using a pre-trained word vector called Template2Vec.

LogRobust~\cite{zhang2019robust} performs supervised binary classification on time series of log messages using an Attention-based Bi-LSTM model.
The method uses a pre-trained Word2Vec model called FastText and combining it with TF-IDF weights to obtain a representation vector of log templates~\cite{goldberg2014word2vec, bojanowski2016enriching}.
By utilizing pre-trained word vectors, LogRobust improves prediction performance for unseen logs.

CNN~\cite{lu2018detecting} is a classification-based log anomaly detection method proposed to detect anomalies in logs by performing time-series analysis with Convolutional Neural Networks~\cite{krizhevsky2017imagenet}.
By using time-series grouped log messages as features, CNN can automatically learn the relationship between log messages and detect anomalies with high accuracy.

Logsy~\cite{nedelkoski2020self} is a parser-free method using the Transformer.
Logsy utilizes anomaly logs generated from another system as auxiliary data for training.
Therefore, although it is a classification-based method, it does not require anomaly data from the monitored system during training.

PLELog~\cite{yang2021semi} is a semi-supervised learning model that incorporates the Positive-Unlabeled (PU) Learning approach~\cite{kiryo2017positive} to log anomaly detection.
This method addresses the problem of insufficient labels using probabilistic label estimation from known normal logs.
The detection model, which consists of an Attention-based Gated Recurrent Unit (GRU) neural network, is trained to classify log sequences into two classes: normal and abnormal.

All of these methods, with the exception of Logsy, analyze the time series of log messages, which causes unnecessary time loss between the actual occurrence of an abnormal log message and its detection.
Since these methods depend on a log parser, they also have limitations in terms of generalization to log messages that are not included in the training data.
Logsy requires anomaly data for training, which comes with limitations for real-world use.
Our proposed method, on the other hand, does not use time-series analysis, which allows for online detection without time lag.
Furthermore, our method does not depend on a log parser or require labeled anomalies in the training data.

\section{Proposed Method}
In this section, we describe our proposed method, LogELECTRA. We also introduce ELECTRA, the method on which LogELECTRA is based.

\subsection{ELECTRA}

\begin{figure*}[htb]
    \centering
    \includegraphics[keepaspectratio, scale=0.5]{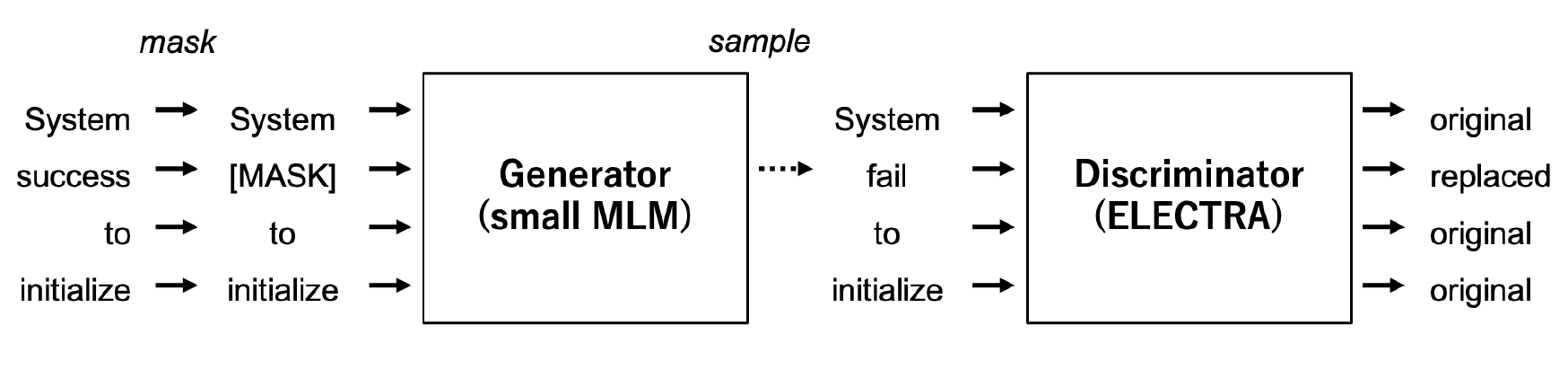}
    \caption{An overview of Replaced Token Detection~\cite{clark2020electra}.}
    \label{fig:electra}
\end{figure*}

ELECTRA is a natural language processing (NLP) model consisting of two Transformers: generator and discriminator)~\cite{clark2020electra, vaswani2017attention}.
ELECTRA is pre-trained by a task called Replaced Token Detection (RTD) as shown in Fig.\ref{fig:electra}.
In this task, the generator replaces tokens in the sequence, and the discriminator attempts to identify which tokens are replaced by the generator in the sequence.
This task replaces Masked Language Modeling (MLM), which is widely used in NLP to train Transformer-based models such as BERT~\cite{devlin2018bert}.
MLM is a self-supervised pre-training objective, masking some tokens in the input sequence and predicting the masked tokens.
Since RTD pre-training makes predictions for all tokens in a sequence, it is sample-efficient compared with MLM pre-training, which only makes predictions for a subset of the tokens in a sequence.
For the same computational complexity, the pre-training model with RTD has been shown to significantly outperform that with MLM in terms of the GLUE score~\cite{wang2018glue}.

The generator is a small Transformer model and is trained by MLM.
It receives masked sequences with some tokens replaced by a special token, [MASK].
The generator attempts to recover the original tokens in the masked index, but since the generator has only a small number of parameters, token recovery is not always successful.
This incorrect restoration of the masked token is used as a token replacement.
Let $\bm{x} = [x_1, x_2, \ldots, x_N]$ be the input sequence of length $N$ and replace $k$ of these tokens with [MASK]. Denoting the masked sequence as $\bm{x^{masked}}$ and the masked indices as $m_i (i = 1 \ldots k)$, the objective function for the generator is as follows.
\begin{align}
    L_g(\bm{x},\bm{\theta_g}) = \mathbb{E} \left( - \sum_{i=1}^k \log p_G(x_{m_i} |\bm{x^{masked}}) \right),
\end{align}
where $\bm{\theta_g}$ is the parameter of generator and  $p_G(x_i|\bm{x^{masked}})$ represents the probability that token $x_i$ is predicted by the generator.

The discriminator is a Transformer model and receives the sequences $\bm{x^{corrupt}}$, where the masked part is replaced by the token predicted by the generator.
Then, for each token, the discriminator predicts whether the token is the original or a replacement.
Let $\bm{h_d(x)}$ be the output of the hidden layer of the discriminator and $\bm{w}$ be the parameter of the sigmoid layer, the objective function for the discriminator is as follows.
\begin{align}
    D(x_i) & = \mathrm{sigmoid}(\bm{w}^T \bm{h_d(x)}_i),
\end{align}
\begin{align}
    L_d(\bm{x},\bm{\theta_d}) & = \mathbb{E} \biggl( \sum_{i=1}^N - \mathbbm{1}(x_i^{corrupt}=x_i) \log (D(x_i^{corrupt})) \nonumber \\
                              & ~~~~~~~~~~ - \mathbbm{1}(x_i^{corrupt}\neq x_i) \log (1 - D(x_i^{corrupt}))  \biggl),
\end{align}
where $\bm{\theta_g}$ is the parameter of discriminator and $\mathbbm{1}$ is an indicator function, a binary function that returns 1 if the condition is satisfied and 0 otherwise.

The final objective function is the sum of the following MLM objective function for the generator and the RTD objective function for the discriminator.
\begin{align}
    \min_{\bm{\theta_g}, \bm{\theta_d}} \sum_{\bm{x} \in \mathcal{X}} L_g(\bm{x},\bm{\theta_g})  + \lambda L_d(\bm{x},\bm{\theta_d}), \label{eq:Loss}
\end{align}
where $\mathcal{X}$ is a large corpus of raw text and $\lambda$ is the hyperparameter of the weight coefficient of the objective functions.
Note that the generator and the discriminator do not compete as in generative adversarial network (GAN), but rather maximize their likelihood~\cite{goodfellow2020generative}.

\subsection{LogELECTRA}
Self-supervised anomaly detection is one of the anomaly detection methods that have attracted attention in recent years and is known to show high performance.
LogELECTRA considers RTD, the ELECTRA pre-training method, as self-supervised learning and utilizes it to detect anomalous logs.
This enables LogELECTRA to deeply analyze a single line of a log message and to detect anomalous log lines as point anomalies with high accuracy.
LogELECTRA proposed in this paper can be divided into three steps: preprocessing, training, and anomaly score calculation.

In the preprocessing step, the standard NLP steps are performed such as lowercasing, removal of stop words, and replacement of structured data such as numbers, dates, IP addresses, and file paths using regular expressions.
Next, we used WordPiece~\cite{wu2016google} to perform tokenization, a method of separating a piece of text into smaller units called tokens.
The tokenizers were trained on a log dataset to ensure that they learn the vocabulary that appears in the log messages of the monitored system.
At the time of testing, we used the tokenizers created in training.
Thus, our method does not use a log parser, the missing information associated with the preprocessing process is minimal, and the detection performance is independent of the performance of the log parser.

In the training step, LogELECTRA learns the same two Transformers as ELECTRA.
The generator and the discriminator are trained by using the normal log dataset after preprocessing.
By using only normal log messages for training, the discriminator can correctly identify the context of normal log messages.
As a result, the discriminator is expected to judge all tokens as original when normal log messages without token replacement are input to the model.
On the other hand, when anomalous log messages are input to the discriminator model, it is expected that some tokens will be judged to have been replaced because anomalous log messages have a different context from normal log messages.
In our proposed method, anomaly detection is based on the difference in judgment of this prediction.

In the step of calculating the anomaly score, the degree of anomaly of the log messages input to the discriminator is calculated.
The anomaly score is designed to be higher the greater the number of tokens judged by the discriminator to have been replaced in the log message.
Let $\bm{x} = [x_1, x_2, \ldots ,x_N]$ for a single line of log messages with length $N$, where $x_i$ is each token in the log message.
The anomaly score for the log message $\bm{x}$ is defined as
\begin{align}
    Score(\bm{x}) = - \frac{1}{N} \sum_{i=1}^N \log (1 - D(x_i)).
\end{align}
This is the average of the objective function of the discriminator if all tokens are original.
This score is expected to be low for normal log messages and high for abnormal log messages.
For each log message, an anomaly score is calculated, and the logs that exceed a certain threshold are marked as abnormal.

\section{Experiments}

\begin{table*}[htb]
    \centering
    \caption{The statistics of datasets}
    \label{tab:dataset}
    \scalebox{1.0}{
        \begin{tabular}{@{}llcccccccc@{}}
            \toprule
            \multicolumn{1}{c}{\multirow{3}{*}{Dataset}}                                                           &
            \multicolumn{1}{c}{\multirow{3}{*}{Grouping}}                                                          &
            \multicolumn{4}{c}{Train}                                                                              &
            \multicolumn{4}{c}{Test}                                                                                 \\ \cmidrule(l){3-10}
            \multicolumn{1}{c}{}                                                                                   &
            \multicolumn{1}{c}{}                                                                                   &
            \multicolumn{2}{c}{\# Log messages}                                                                    &
            \multicolumn{2}{c}{\begin{tabular}[c]{@{}c@{}}\# Log sequences \\ (random/chronological)\end{tabular}} &
            \multicolumn{2}{c}{\# Log messages}                                                                    &
            \multicolumn{2}{c}{\begin{tabular}[c]{@{}c@{}}\# Log sequences \\ (random/chronological)\end{tabular}}   \\ \cmidrule(l){3-10}
            \multicolumn{1}{c}{}                                                                                   &
            \multicolumn{1}{c}{}                                                                                   &
            Normal                                                                                                 &
            Anomaly                                                                                                &
            Normal                                                                                                 &
            Anomaly                                                                                                &
            Normal                                                                                                 &
            Anomaly                                                                                                &
            Normal                                                                                                 &
            Anomaly                                                                                                  \\ \midrule
            \multirow{2}{*}{BGL}                                                                                   &
            60 minutes                                                                                             &
            3059327                                                                                                &
            1120874                                                                                                &
            2884 / 2625                                                                                            &
            536 / 496                                                                                              &
            266085                                                                                                 &
            267207                                                                                                 &
            722 / 981                                                                                              &
            129 / 171                                                                                                \\ \cmidrule(l){2-10}
                                                                                                                   &
            100 messages                                                                                           &
            3362700                                                                                                &
            408100                                                                                                 &
            37708 / 55401                                                                                          &
            4081 / 25066                                                                                           &
            858300                                                                                                 &
            84393                                                                                                  &
            8583 / 13851                                                                                           &
            844 / 6309                                                                                               \\ \midrule
            \multirow{2}{*}{Spirit}                                                                                &
            60 minutes                                                                                             &
            329396                                                                                                 &
            3359245                                                                                                &
            2884 / 2625                                                                                            &
            536 / 496                                                                                              &
            615334                                                                                                 &
            696025                                                                                                 &
            722 / 981                                                                                              &
            129 / 171                                                                                                \\ \cmidrule(l){2-10}
                                                                                                                   &
            100 messages                                                                                           &
            2061600                                                                                                &
            1938400                                                                                                &
            37708 / 55401                                                                                          &
            4009 / 25066                                                                                           &
            965400                                                                                                 &
            34600                                                                                                  &
            9427 / 13851                                                                                           &
            817 / 6309                                                                                               \\ \midrule
            \multirow{2}{*}{Thunderbird}                                                                           &
            60 minutes                                                                                             &
            2070050                                                                                                &
            7081597                                                                                                &
            2884 / 2625                                                                                            &
            536 / 496                                                                                              &
            502107                                                                                                 &
            346246                                                                                                 &
            722 / 981                                                                                              &
            129 / 171                                                                                                \\ \cmidrule(l){2-10}
                                                                                                                   &
            100 messages                                                                                           &
            4286900                                                                                                &
            3713100                                                                                                &
            37708 / 55401                                                                                          &
            4009 / 25066                                                                                           &
            1807800                                                                                                &
            192200                                                                                                 &
            9427 / 13851                                                                                           &
            817 / 6309                                                                                               \\ \bottomrule
        \end{tabular}
    }
\end{table*}

In this paper, the experiment was conducted in accordance with the survey paper~\cite{le2022log} to compare the performance of our method with existing methods.
The following describes the dataset used in our experiments, the comparison methods, the implementation details for our method, evaluation metrics, and the results.

\subsection{Dataset}
Three public datasets (BGL, Spirit, and Thunderbird) were used to evaluate the model of log anomaly detection~\cite{he2020loghub,oliner2007supercomputers}.
Details of each dataset are as follows.

\begin{itemize}
    \item BGL (Blue Gene/L) is a log dataset of supercomputing systems collected by Lawrence Livermore National Labs.
          Each message in this dataset was manually labeled as either normal or abnormal.
    \item Spirit is a collection of system log data from the Spirit supercomputing system installed at Sandia National Labs (SNL).
          Since this dataset contains a large number of log messages, we use a small set containing the first 5 million log lines of the original Spirit dataset in this experiments.
    \item Thunderbird dataset is a log datasset collected from SNL's Thunderbird supercomputer.
          Since the Thunderbird is also very large, we use a small set of the first 10 million in this experiments.
\end{itemize}

\subsection{Comparison methods}

We compared our proposed method LogELECTRA with DeepLog, LogAnomaly, PLELog, LogRobust, and CNN~\cite{du2017deeplog, meng2019loganomaly, zhang2019robust, lu2018detecting ,yang2021semi}.
These methods have publicly available source code; we did not include Logsy~\cite{nedelkoski2020self} in the comparison since its source code was not publicly available.
For hyperparameters, the same values were used for models reported in the original papers; otherwise, hyperparameters were adjusted empirically.
For the log parser, we used Drain~\cite{he2017drain} to split the logs into log templates and log parameters for all models requiring parsing.

In addition, LogAnomaly is trained by adding domain-specific antonyms and synonyms to the template2vec model in the original paper, but this information was not available, so we used FastText~\cite{bojanowski2016enriching}, a pre-trained Word2Vec, to compute the semantic vector of the log templates.
For PLELog, we used a Glove~\cite{pennington2014glove}, another pre-trained word vector model, to compute the semantic vectors of the log templates, as in the original paper~\cite{yang2021semi}.

These five existing methods detect anomalies on the basis of the time series of log messages unlike our method.
Therefore, for these methods,multiple log messages need to be grouped in some way before training.
In this experiments, log messages are grouped into log sequences by using a fixed window strategy that groups them by fixed width in accordance with the timestamp of occurrence.
Also, in accordance with the ground truth labels of each datasets, a log sequence is considered anomalous if it contains an anomalous log message.
If all log messages contained in the log sequence are normal, the log sequence is considered normal.
In this paper, experiments were conducted for two patterns of fixed widths: 100 messages and 60 minutes.

On the other hand, unlike other existing methods, LogELECTRA performs anomaly detection on a single line of log messages.
Therefore, in this experiment, to compare its detection performance with those of existing methods, a log sequence consisting of multiple log messages was input to LogELECTRA and the detection performance was evaluated.
If there was even one abnormal log message in the log sequence, the log sequence was marked as abnormal by LogELECTRA.

The method of selecting training data also affects the accuracy of anomaly detection.
There are two methods for selecting training data: chronological selection and random selection.
Since the state of the monitored system usually changes with time, log messages that are not included in the training data may appear in the test data in a chronological selection~\cite{7476654}.
On the other hand, in random selection, log messages are randomly shuffled and then split into training data and test data.
In this way, the training data contains more diverse information.
Although random selection is suitable for measuring the performance of the detection method itself, it is not a realistic scenario that can be used in a real environment.
In this paper we experimented under both selection methods.
The split ratio was set as train : test = 80 : 20.

Table \ref{tab:dataset} summarizes the details of the dataset after each grouping under each selection method.

\begin{table*}[htb]
    \centering
    \caption{Comparison of detection performance of each model on BGL dataset.}
    \label{tab:bgl}
    \scalebox{1.0}{
        \begin{tabular}{@{}lllllllll@{}}
            \toprule
            \multicolumn{1}{c}{\multirow{2}{*}{Model}}                     &
            \multicolumn{4}{c}{60 minutes grouping (random/chronological)} &
            \multicolumn{4}{c}{100 messages grouping (random/chronological)}                                                                                                                               \\ \cmidrule(l){2-9}
            \multicolumn{1}{c}{}                                           &
            \multicolumn{1}{c}{Prec}                                       &
            \multicolumn{1}{c}{Rec}                                        &
            \multicolumn{1}{c}{Spec}                                       &
            \multicolumn{1}{c}{F1}                                         &
            \multicolumn{1}{c}{Prec}                                       &
            \multicolumn{1}{c}{Rec}                                        &
            \multicolumn{1}{c}{Spec}                                       &
            \multicolumn{1}{c}{F1}                                                                                                                                                                         \\ \midrule
            LogELECTRA                                                     & 0.845 / 0.882 & 0.955 / 0.882 & 0.943 / 0.968 & 0.897 / 0.882 & 0.945 / 0.523 & 0.983 / 0.946 & 0.993 / 0.706 & 0.963 / 0.674 \\
            DeepLog                                                        & 0.926 / 0.311 & 0.773 / 0.941 & 0.982 / 0.440 & 0.842 / 0.468 & 0.972 / 0.298 & 0.737 / 0.338 & 0.997 / 0.921 & 0.838 / 0.317 \\
            LogAnomaly                                                     & 0.915 / 0.325 & 0.730 / 0.811 & 0.980 / 0.547 & 0.812 / 0.465 & 0.921 / 0.187 & 0.694 / 0.915 & 0.992 / 0.610 & 0.792 / 0.312 \\
            PLELog                                                         & 0.965 / 0.802 & 0.987 / 0.625 & 0.996 / 0.943 & 0.976 / 0.701 & 0.646 / 0.925 & 0.856 / 0.402 & 0.959 / 0.985 & 0.736 / 0.560 \\ \midrule
            RobustLog                                                      & 0.993 / 0.973 & 0.957 / 0.961 & 0.998 / 0.993 & 0.975 / 0.967 & 0.974 / 0.906 & 0.985 / 0.951 & 0.996 / 0.991 & 0.979 / 0.928 \\
            CNN                                                            & 0.985 / 0.921 & 0.852 / 0.908 & 0.996 / 0.978 & 0.914 / 0.914 & 0.971 / 0.953 & 0.982 / 0.949 & 0.996 / 0.995 & 0.976 / 0.951 \\ \bottomrule
        \end{tabular}
    }
\end{table*}

\begin{table*}[htb]
    \centering
    \caption{Comparison of detection performance of each model on Spirit dataset.}
    \label{tab:spirit}
    \scalebox{1.0}{
        \begin{tabular}{@{}lllllllll@{}}
            \toprule
            \multicolumn{1}{c}{\multirow{2}{*}{Model}}                     &
            \multicolumn{4}{c}{60 minutes grouping (random/chronological)} &
            \multicolumn{4}{c}{100 messages grouping (random/chronological)}                                                                                                                               \\ \cmidrule(l){2-9}
            \multicolumn{1}{c}{}                                           &
            \multicolumn{1}{c}{Prec}                                       &
            \multicolumn{1}{c}{Rec}                                        &
            \multicolumn{1}{c}{Spec}                                       &
            \multicolumn{1}{c}{F1}                                         &
            \multicolumn{1}{c}{Prec}                                       &
            \multicolumn{1}{c}{Rec}                                        &
            \multicolumn{1}{c}{Spec}                                       &
            \multicolumn{1}{c}{F1}                                                                                                                                                                         \\ \midrule
            LogELECTRA                                                     & 0.879 / 0.961 & 1.000 / 0.943 & 0.447 / 0.968 & 0.935 / 0.952 & 0.928 / 0.687 & 0.994 / 0.936 & 0.949 / 0.977 & 0.960 / 0.792 \\
            DeepLog                                                        & 0.811 / 0.457 & 1.000 / 1.000 & 0.063 / 0.008 & 0.896 / 0.627 & 0.769 / 0.271 & 0.891 / 0.739 & 0.823 / 0.653 & 0.826 / 0.397 \\
            LogAnomaly                                                     & 0.803 / 0.462 & 1.000 / 1.000 & 0.021 / 0.011 & 0.891 / 0.632 & 0.797 / 0.296 & 0.899 / 0.751 & 0.849 / 0.629 & 0.845 / 0.425 \\
            PLELog                                                         & 0.909 / 0.952 & 0.918 / 0.507 & 0.660 / 0.861 & 0.914 / 0.662 & 0.865 / 0.199 & 0.807 / 0.654 & 0.910 / 0.970 & 0.867 / 0.305 \\ \midrule
            RobustLog                                                      & 0.999 / 0.991 & 1.000 / 1.000 & 0.999 / 0.992 & 0.999 / 0.995 & 0.999 / 0.928 & 0.998 / 0.823 & 0.999 / 0.997 & 0.998 / 0.872 \\
            CNN                                                            & 0.999 / 0.981 & 1.000 / 0.926 & 0.999 / 0.983 & 0.999 / 0.971 & 0.999 / 0.757 & 0.998 / 0.838 & 0.999 / 0.991 & 0.998 / 0.796 \\ \bottomrule
        \end{tabular}
    }
\end{table*}

\begin{table*}[htb]
    \centering
    \caption{Comparison of detection performance of each model on Thunderbird dataset.}
    \label{tab:thunderbird}
    \scalebox{1.0}{
        \begin{tabular}{@{}lllllllll@{}}
            \toprule
            \multicolumn{1}{c}{\multirow{2}{*}{Model}}                     &
            \multicolumn{4}{c}{60 minutes grouping (random/chronological)} &
            \multicolumn{4}{c}{100 messages grouping (random/chronological)}                                                                                                                               \\ \cmidrule(l){2-9}
            \multicolumn{1}{c}{}                                           &
            \multicolumn{1}{c}{Prec}                                       &
            \multicolumn{1}{c}{Rec}                                        &
            \multicolumn{1}{c}{Spec}                                       &
            \multicolumn{1}{c}{F1}                                         &
            \multicolumn{1}{c}{Prec}                                       &
            \multicolumn{1}{c}{Rec}                                        &
            \multicolumn{1}{c}{Spec}                                       &
            \multicolumn{1}{c}{F1}                                                                                                                                                                         \\ \midrule
            LogELECTRA                                                     & 0.778 / 0.723 & 1.000 / 0.838 & 0.664 / 0.901 & 0.875 / 0.776 & 0.967 / 0.643 & 0.995 / 0.997 & 0.978 / 0.922 & 0.981 / 0.781 \\
            DeepLog                                                        & 0.694 / 0.535 & 1.000 / 0.967 & 0.601 / 0.648 & 0.819 / 0.689 & 0.926 / 0.645 & 0.872 / 0.942 & 0.956 / 0.945 & 0.898 / 0.766 \\
            LogAnomaly                                                     & 0.625 / 0.547 & 1.000 / 0.935 & 0.454 / 0.676 & 0.771 / 0.691 & 0.924 / 0.586 & 0.891 / 0.936 & 0.954 / 0.927 & 0.907 / 0.729 \\
            PLELog                                                         & 0.960 / 0.903 & 0.676 / 0.667 & 0.941 / 0.952 & 0.793 / 0.767 & 0.965 / 0.958 & 0.961 / 0.624 & 0.983 / 0.993 & 0.967 / 0.756 \\ \midrule
            RobustLog                                                      & 1.000 / 0.960 & 0.920 / 0.774 & 1.000 / 0.986 & 0.958 / 0.857 & 0.999 / 0.982 & 0.993 / 0.705 & 0.999 / 0.998 & 0.996 / 0.705 \\
            CNN                                                            & 0.998 / 0.958 & 0.932 / 0.741 & 1.000 / 0.986 & 0.963 / 0.836 & 0.996 / 0.919 & 0.992 / 0.864 & 0.999 / 0.991 & 0.995 / 0.891 \\ \bottomrule
        \end{tabular}
    }
\end{table*}

\subsection{Implementation details}

We describe the hyper-parameters of LogELECTRA. As a discriminator, we used a Transformer model with 512 dimensional 6 hidden layers. As the generator, we used a model with three 256-dimensional hidden layers. The mask probability of the input to the generator was set to 15$\%$ and $\lambda = 50$ was used as the weight coefficient for the objective function in Eq.\ref{eq:Loss}. AdamW was used as the optimizer~\cite{loshchilov2017decoupled}. Other parameters were tuned empirically.

\subsection{Evaluation Metrics}
To measure the effectiveness of the models in detecting anomalies, we use the Precision, Recall, Specificity, and F1-Score metrics defined as follows.
\begin{itemize}
    \item Precision is the percentage of logs detected as anomalies in the model that were truly anomalous. $\mathrm{Prec} =\frac{\mathrm{\mathrm{TP}}}{\mathrm{TP} + \mathrm{FP}}$.
    \item Recall is the percentage of anomaly logs detected by the model out of all true anomaly logs. $\mathrm{Rec} = \frac{\mathrm{TP}}{\mathrm{TP} + \mathrm{FN}}$.
    \item Specificity is the percentage of logs identified as normal by the model out of all true normal logs. $\mathrm{Spec} = \frac{\mathrm{TN}}{\mathrm{TN} + \mathrm{FP}}$.
    \item F1 score is The harmonic mean of Precision and Recall. $\mathrm{F1} = \frac{2 * \mathrm{Prec} * \mathrm{Rec}}{\mathrm{Prec} + \mathrm{Rec}}$.
\end{itemize}
TP (True Positive) is the number of abnormal log sequences correctly detected by the model;
FP (False Positive) is the number of normal log sequences falsely identified as anomaly;
FN (False Negative) is the number of abnormal log sequences not detected by the model;
and TN (True Negative) is the number of normal log sequences not detected by the model.
For methods that require threshold adjustment, including our method, the optimal threshold was determined by F1 maximization.

\subsection{Results}
Comparison results of detection performance on the BGL, Spirit, and Thunderbird datasets are shown in tables \ref{tab:bgl}, \ref{tab:spirit}, and \ref{tab:thunderbird}, respectively.
These results show that our LogELECTRA achieved higher performance in all experimental settings than other unsupervised methods such as DeepLog and LogAnomaly.
They also show that LogELECTRA performs comparably to or better than semi-supervised methods such as PLELog.
In particular, under the chronological training data selection, LogELECTRA has higher detection performance than other methods that do not use anomaly labels for training.
Chronological training data selection is a realistic strategy.
Furthermore, in some experiments, LogELECTRA performs comparably to supervised methods, despite the unfair settings.

Existing unsupervised and semi-supervised methods depend on the grouping method for detection performance.
This is especially true for the Spirit dataset, where the F1 Scores for DeepLog and LogAnomaly are significantly degraded more in the 100-messages grouping setting than in the 60-min grouping setting.
On the other hand, LogELECTRA is able to achieve relatively stable and high performance with any grouping method.
This shows that LogELECTRA is less dependent on the grouping method, since it specializes in the analysis of a single log message.
The results indicate that LogELECTRA can quickly find anomalous logs even under diverse circumstances.

When training data is selected for chronological, the detection performance of the existing unsupervised and semi-supervised methods is degraded compared with that of randomly selected training data.
This indicates that chronological train data selection is a more difficult setting than random train data selection.
The reason is that with random selection, the model can make more accurate predictions since it can see future log events during the training phase.
On the other hand, LogELECTRA's accuracy degradation is less than that of other unsupervised and semi-supervised methods, even in the chronological setting.
This indicates that LogELECTRA generalizes well to unseen normal logs and is able to capture the context of normal logs differently from anomaly logs.

Focusing on supervised methods, even in the case of chronological setting, the detection performance is not significantly different from that of random data selection.
This suggests that if the features that separate normal and abnormal contexts are captured, high accuracy can be achieved even in the case of chronological data selection, even with unlabeled learning.
Future challenges include capturing this feature from normal log messages alone, without label information, and closing the gap between LogELECTRA and these supervised methods.

\section{Conclusion}
In this paper, we proposed a novel log anomaly detection method, LogELECTRA.
Our method utilizes the pre-training task of ELECTRA, Replaced Token Detection, to capture the context of normal log messages and detect anomalous log messages with a different context.
LogELECTRA is specialized to detect point anomalies, so there is no unnecessary delay in detection unlike in existing methods.
Furthermore, since our method is based on the ELECTRA, a superior natural language processing model, it accurately captures the context of normal log messages and has the advantage of generalization performance for unseen log messages.
This was confirmed in experiments on three different public log datasets.
LogELECTRA outperforms existing state-of-the-art methods that do not use anomaly data for training and achieves detection performance comparable to supervised methods under some settings.
Future work will include studies for real-world adaptations, such as scalability, additional experiments on other system logs, model transferability, and measurement of actual anomaly-to-detection delays.

\bibliographystyle{IEEEtran}
\bibliography{IEEEabrv,main}

\end{document}